\pdfoutput=1

\documentclass[11pt]{article}

\usepackage[]{naacl2021}

\usepackage{times}
\usepackage{latexsym}

\usepackage[T1]{fontenc}

\usepackage[utf8]{inputenc}

\usepackage{microtype}

\usepackage{tabularx}
\usepackage{multirow, makecell}
\newcolumntype{R}[1]{>{\hsize=#1\hsize\raggedleft\arraybackslash}X}%
\newcolumntype{L}[1]{>{\hsize=#1\hsize\raggedright\arraybackslash}X}%
\newcolumntype{C}[1]{>{\hsize=#1\hsize\centering\arraybackslash}X}%

\usepackage{algorithmicx}
\usepackage{algpseudocode}
\usepackage{graphics}
\usepackage{graphicx}

\newcommand{\comment}[1]{}

\newcommand{\tekgen}{\textsc{TeKGen}}
\newcommand{\kelm}{\textsc{KeLM}}

\usepackage{tikz}
\newcommand*\circled[1]{\tikz[baseline=(char.base)]{\node[shape=circle,draw,inner sep=1pt] (char) {#1};}}

\title{Knowledge Graph Based Synthetic Corpus Generation for Knowledge-Enhanced Language Model Pre-training}

\author{Oshin Agarwal\thanks{\enskip Work done during internship at Google}\hspace{0.3em}\textsuperscript{1} \quad Heming Ge\textsuperscript{2} \quad Siamak Shakeri\textsuperscript{2} \quad Rami Al-Rfou\textsuperscript{2} \\
  \textsuperscript{1}{ University of Pennsylvania} \quad \textsuperscript{2}{Google Research} \\
  \texttt{\small oagarwal@seas.upenn.edu, \{hemingge, siamaks, rmyeid\}@google.com}}

\date{}

\begin{document}
\maketitle

\begin{abstract}

Prior work on Data-To-Text Generation, the task of converting knowledge graph (KG) triples into natural text, focused on domain-specific benchmark datasets. 
In this paper, however, we verbalize the entire English Wikidata KG, and discuss the unique challenges associated with a broad, open-domain, large-scale verbalization. 
We further show that verbalizing a comprehensive, encyclopedic KG like Wikidata can be used to integrate structured KGs and natural language corpora. 
In contrast to the many architectures that have been developed to integrate these two sources, our approach converts the KG into natural text, allowing it to be seamlessly integrated into existing language models. 
It carries the further advantages of improved factual accuracy and reduced toxicity in the resulting language model. 
We evaluate this approach by augmenting the retrieval corpus in a retrieval language model and showing significant improvements on the knowledge intensive tasks of open domain QA and the LAMA knowledge probe.
\end{abstract}

\section{Introduction}

\begin{figure}
    \centering
    \includegraphics[scale=0.45,trim=3cm 7cm 3cm 1cm,clip]{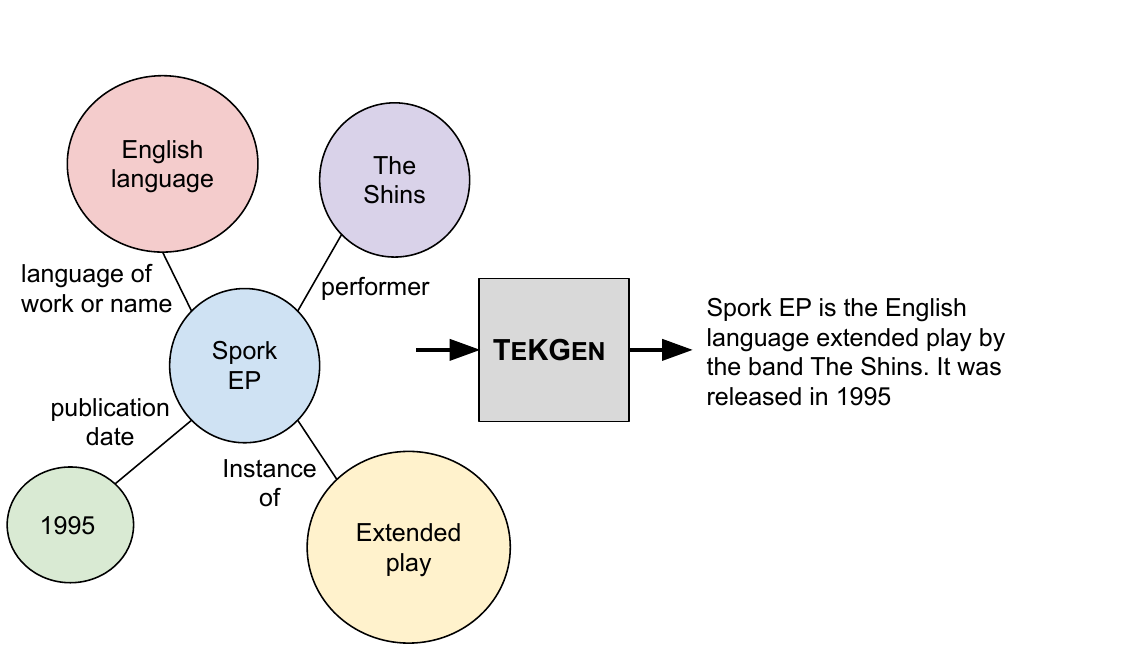}
    \caption{An example of generating text from KG. First, the entity subgraphs on the left are created and then converted to the sentence on the right.}
  \label{fig:verbalization}
\end{figure}

Data-To-Text Generation \cite{kukich-1983-design, 294135} involves converting knowledge graph (KG) triples of the form \texttt{\small (subject, relation, object)} into a natural language sentence(s). There are many standard datasets for this task such as WebNLG \cite{gardent-etal-2017-webnlg} and many systems have been developed to improve performance on these datasets. However, to the best of our knowledge, no prior work has attempted to verbalize a full knowledge graph. Verbalizing a full KG has additional challenges over small benchmark datasets, such as entity and relation coverage and the lack of grouped sets of triples that can produce coherent sentences together. In this paper, we convert the {\em English} Wikidata KG \cite{42240} into natural language text (Figure \ref{fig:verbalization}). The generated corpus, which we call the \kelm{} Corpus, consists of $\sim$18M sentences spanning $\sim$45M triples with $\sim$1500 distinct relations. For training the verbalization system, we also create an {\em English} Wikidata KG--Wikipedia Text aligned corpus consisting of a variety of entities such as dates and numerical quantities.

\begin{figure*}
    \centering
    \includegraphics[width=0.95\textwidth,trim=0cm 1.5cm 0cm 0cm,clip]{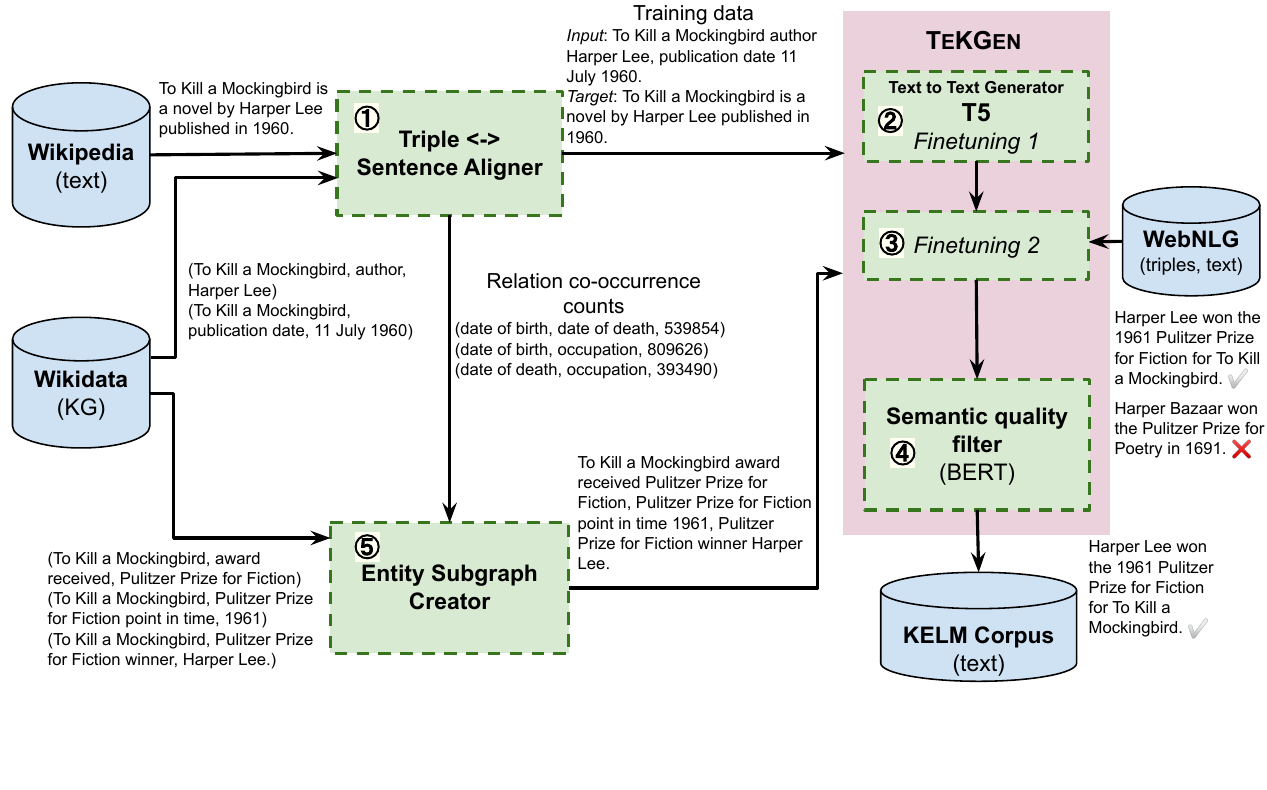}
    \caption{Pipelines for training the \tekgen{} model and generating the \kelm{} corpus.
    In Step \circled{1}, KG triples are aligned with Wikipedia text using distant supervision. In Steps \circled{2} \& \circled{3}, T5 is fine-tuned sequentially first on this corpus, followed by a small number of steps on the WebNLG corpus, 
    In Step \circled{4}, BERT is fine-tuned to generate a semantic quality score for generated sentences w.r.t. triples. Steps \circled{2}, \circled{3} \& \circled{4} together form \tekgen{}.
    To generate the \kelm{} corpus, 
     in Step \circled{5}, entity subgraphs are created using the relation pair alignment counts from the training corpus generated in step \circled{1}.
     The subgraph triples are then converted into natural text using \tekgen{}.}
  \label{fig:model}
\end{figure*}

We evaluate the quality of the generated corpus through human evaluation of a random sample. We further showcase the utility of this corpus in language model pre-training. Text represents a limited coverage of the world knowledge. Therefore, we expect the language models to be restricted to facts that are expressed in natural language. Moreover, facts may not be expressed as explicitly in text as they are in KGs, and the variability in the quality of text can eventually cause biases in the resulting models \cite{bolukbasi2016man, sheng-etal-2019-woman, manzini-etal-2019-black}. Building models that handle structured data and free form text seamlessly has been a long sought-after goal. However, their integration is challenging due to different structural formats. KG verbalization provides a simple way to integrate KGs with natural text. We illustrate this by augmenting the REALM \cite{guu2020realm} retrieval corpus with the \kelm{} Corpus. We evaluate the augmented system on the LAMA knowledge probe  \cite{petroni-etal-2019-language} and open domain QA and show improvements on both. Through ablation experiments where we augment the retrieval corpus with the raw triples instead, we further confirm the effectiveness of verbalization.
Our contributions are as follows - 

\begin{itemize}
    \setlength\itemsep{0em}
    \item {\tekgen} ({\bf Te}xt from {\bf KG Gen}erator): A data-to-text sequence-to-sequence model for verbalizing an entire KG 
    \item {\tekgen} training corpus: Text--KG aligned corpora with a wide variety of relations including dates and quantities 
    \item \kelm{} Corpus, \footnote{Both the \tekgen{} training corpus and the \kelm{} corpus are available at \url{https://github.com/google-research-datasets/KELM-corpus}} (Corpus for {\bf K}nowledge-{\bf E}nhanced {\bf L}anguage {\bf M}odel Pre-training): A large-scale synthetic corpus of Wikidata KG as natural text
    \item Data-to-text generation as a method to integrate KGs with textual pre-training corpora, showing improvements on open domain QA and LAMA probe with the augmented model
\end{itemize}

\section{{\tekgen}}

One of the challenges in converting an entire KG to text is the wide variety of entities and relations. Wikidata consists of $\sim$6M entities and $\sim$1500 relations. In comparison, the WebNLG dataset has $\sim$600 entities and $\sim$20 relations. In this section, we discuss the various components of \tekgen{}, also illustrated in Figure \ref{fig:model} -- 

\begin{enumerate}
    \itemsep0em 
    \item Create a large yet noisy training dataset using distant supervision.
    \item Sequentially fine-tune T5, first on the dataset from step 1 for improved coverage, then on a small clean dataset for reduced hallucination.
    \item Build a filter for the generated text based on its semantic quality w.r.t. the KG triples.
\end{enumerate}

\subsection{Training Dataset}

We first create training data using distant supervision by aligning Wikidata triples to Wikipedia text (see Figure \ref{fig:alignment_algo}).

\subsubsection{KG--Text Alignment}
\label{sec:alignment}
For each entity, we constrain the candidate sentences to the root section of its Wikipedia page because this section generally describes the relations of the subject entity with other entities. For each sentence in this section, we match all triples that have this entity as the subject. A triple is said to match if any alias of the object entity occurs in the sentence. We do not match relations to text as there are too many ways to express them. Constraining to the subject entity's page and root section generally ensures that the relation is expressed in the sentence if it mentions the object entity. Each triple can align to multiple sentences and each sentence can have multiple triples aligned to it. If any alias of the subject entity occurs in the given sentence, the sentence is selected as is, else the first animate third-person personal or possessive pronoun is replaced by the subject entity's canonical name. The pronoun replacement heuristic also works well because of this constraint. All triples aligned to a given sentences are combined together as a single example.

\begin{figure}[t]
\fbox{
\noindent
\hspace{-1.5em}
\begin{minipage}{\columnwidth}
    \begin{algorithmic}
    \small
    \State $\textrm{alignment\_pairs}\gets \{\}$
    \ForAll{$\textrm{sentences } t \in \textrm{root section of Wiki page of entity } s$}
    \ForAll{$\textrm{triples } (s, r, o) \in KG$}
    \If{$t\textrm{.contains}(\textrm{alias}(o))$}
    \If{$t\textrm{.notcontains}(\textrm{alias}(s))$}
    \State $p\gets t.\textrm{first\_pronoun}$
    \State $t\gets t\textrm{.replace}(p, \textrm{name}(s))$
    \EndIf
    \State $\textrm{alignment\_pairs.add}((t, (s, r, o)))$
    \EndIf
    \EndFor
    \EndFor
    \end{algorithmic}
\end{minipage}
}
\caption{KG--Text alignment algorithm.}
\label{fig:alignment_algo}
\end{figure}

Alignment statistics are shown in Table \ref{table:alignStats} and some alignment examples are shown in Table \ref{table:KalignEx}. There are a total of $\sim$45M triples, $\sim$35\% of which were aligned to sentences. This results in $\sim$8M examples, covering $\sim$42\% of the relations.

Note that each sentence in the aligned corpus is matched to triples with a common subject entity. While this results in some noise, such errors should be small due to the constraint that the text is the root section of the subject entity page. This constraint allows us to maintain the same property of common subject entity as the entity subgraph used in inference (\S \ref{sec:kelm_corpus}). It also simplified the alignment process, removing the need to match relations to text. In comparison, the T-REx \cite{elsahar-etal-2018-rex} corpus does not have this noise due the use of typical NLP pipeline with coreference resolution and predicate linking. However, it suffers from errors due to entity linking and incorrect entailment, which are unlikely in our corpus due to this constraint. 


\begin{table}[t]
\centering
\small
\setlength{\tabcolsep}{4pt}
\begin{tabularx}{0.7\linewidth}{L{0.7}R{0.3}}

Total KG triples & 45,578,261\\
Triples aligned & 16,090,457\\
Total sentences aligned & 7,978,814\\
Total KG relations & 1,522\\
Relations aligned & 663\\
\hline
\end{tabularx} 
\caption{KG--Text alignment statistics.}
\label{table:alignStats}
\end{table}

\begin{table*}[t]
\centering
\footnotesize
\setlength{\tabcolsep}{4pt}
\begin{tabularx}{\linewidth}{L{0.56}|L{0.52}}
\multicolumn{1}{c|}{\bf Input Triples} & \multicolumn{1}{c}{\bf Target Sentence} \\\hline
Das Tagebuch der Anne Frank, (distributor, Universal Pictures), (country, Germany), (publication date, 03 March 2016) & The film was theatrically released in the Germany on March 3, 2016, by Universal Pictures International. \\\hline
Neff Maiava, (date of birth, 01 May 1924), (date of death, 21 April 2018), (occupation, professional wrestler) & Maiava (May 1, 1924 April 21, 2018) was an American Samoan professional wrestler. \\\hline
Barack Obama 2012 presidential campaign, (country, United States), (end time, 06 November 2012), (start time, 04 April 2011) & The 2012 reelection campaign of Barack Obama, the 44th President of the United States, was formally announced on April 4, 2011.\\\hline
Blue whale (parent taxon, Balaenoptera) & The blue whale (Balaenoptera musculus) is a marine mammal belonging to the baleen whale suborder Mysticeti. \\
\end{tabularx} 
\caption{Examples of alignment (training data).}
\label{table:KalignEx}
\end{table*}

\subsubsection{Types of Triples}
\label{sec:triple_types}
We extract several types of triples, each of which have slightly different matching techniques. Other alignment corpora built using Wikipedia hyperlinks \cite{chen2020kgpt, logan-etal-2019-baracks} would miss many of these triples with entities without Wikipedia pages such as quantities, dates and certain occupations, and hence relations such as date of birth, publication year and distance from Earth.

\begin{enumerate}
    \itemsep0em 
    \item Object entity with a Wikipedia page: These are aligned by string matching all aliases. (e.g.\ Barack Obama)
    \item Object entity without a Wikipedia page: These are also aligned by matching all aliases. (e.g.\ skateboarder, professional wrestler)
    \item Object entity is a quantity: They have two components -- Amount and Units. Units are also entities in the KG and have aliases. We concatenate the amount with each of the unit's aliases for matching (e.g.\ 16 km/hr, 16 km/h, 16 kilometres per hour). Certain quantities do not have units (e.g.\ When the relation is number of episodes).
    \item Object entity is a date: Wikipedia uses only three date formats. \footnote{\url{https://en.wikipedia.org/wiki/Wikipedia:Date_formattings}} We first find all dates in a sentence using regular expressions and parse them into a structured format containing day of the month, month and year. If any of these components exist in both the dates to be matched, they should match. For example, if the triple date has all three components but the date extracted from a sentence has only the year, then only the year needs to match.  
    \item Relations with a subproperty: For instance, the relation {\em award received} has the subproperty {\em year} and the relation {\em spouse} may have the subproperty {\em place of marriage}. 
    We retain the main triple as such and reformat the subproperty as a triple of the form \texttt{\small (subject\_entity, object\_entity subproperty\_name, subproperty\_value)} e.g.\ (Barack, spouse, Michelle) has the subproperty (place of marriage, Trinity Church). These are converted to (Barack, spouse, Michelle) and (Barack, Michelle place of marriage, Trinity Church).
\end{enumerate}

While the type of the triples is important in the alignment process, the verbalization model is agnostic to the type and treats all triples the same. 

\subsection{Model}
\label{sec:gen_model}
We perform a two-step sequential finetuning of the pre-trained T5-large \cite{raffel2020exploring} model for converting triples to text. Triples are concatenated as \texttt{\small subject relation\_1 object\_1, ....relation\_n object\_n} for input to T5. The model is first fine-tuned on the aligned corpus for 5000 steps to increase the coverage of entities and relations. However, this results in the generation of Wikipedia-like sentences and hallucination if a certain expected input triple is missing. For example, Wikipedia sentences generally mention {\em date of birth}, {\em date of death}, {\em occupation} together. If the occupation is missing in the input, the system hallucinates a random occupation. ``Neff Maiava date of birth 01 May 1924, date of death, 21 April 2018.'' generates ``Neff Maiava (1 May 1924 - 21 April 2018) was an Albanian actor.''; hallucinating a profession. To overcome this, we further fine-tune the model on WebNLG 2017 data for 500 steps. While WebNLG has low coverage, the information in the input triples matches the target sentence exactly. WebNLG also has a different sentence structure than Wikipedia. This reduces conformity to Wikipedia sentence structure and hence reduces hallucination. We use a learning rate of 0.001, a batch size of 1048576 tokens and a maximum decoding length of 256. 

\begin{table}[t]
\centering
\small
\setlength{\tabcolsep}{4pt}
\begin{tabularx}{0.6\linewidth}{L{0.7}R{0.2}}

Pearson correlation & 0.73 \\
Spearman correlation & 0.66 \\
Kendall's Tau & 0.51 \\
\hline
\end{tabularx} 
\caption{Semantic Filtering Evaluation.}
\label{table:FilteringEval}
\end{table}

\subsection{Quality Filtering}
\label{sec:filtering}

We perform a semantic quality based filtering of the sentences generated by the triple-to-text module. This is a separate post-processing module used during inference and is not jointly optimized with the text generation module. A semantic quality score is assigned to each generated sentence w.r.t.\ the input triples that indicates whether or not the generated text captures the full meaning of the triples and does not hallucinate extra information. The score is generated using a BERT base uncased model with input of the form \texttt{\small [CLS] concatenated-triples [SEP] reference-or-generated-sentence}. It is fine-tuned for 1000 steps on the WebNLG 2017 human assessment data. The data contains system predictions submitted to the shared task rated on a scale of 1-3 for semantics and fluency. We use the semantics score and scale it to 0-1. We also add gold references with a score of 1. This results in 2706 examples, 90\% of which are used for finetuning and the remaining for evaluation. High correlations are obtained between the predicted scores and human scores on the evaluation split (Table \ref{table:FilteringEval}). 

\section{\kelm{} Corpus}
\label{sec:kelm_corpus}
In this section, we utilize the \tekgen{} model and filtering mechanism to build a synthetic corpus that captures the KG in natural language format.

\subsection{Entity Subgraph}
Datasets such as WebNLG have instances with grouped triples that can be expressed as a fluent sentence. Such groups are not available for a large KG and using one triple at a time for inference would lead to hallucination as training uses multiple triples per example. Therefore, we develop a strategy to create entity subgraphs based on relation co-occurrence counts i.e. frequency of alignment of two relations to the same sentence in the training data. 
The algorithm is shown in Figure \ref{fig:grouping_algo}. 
It produces $\sim$18M entity subgraphs from $\sim$45M triples so the final corpus will have 18M generated sentences corresponding to each entity subgraph.

\begin{figure}[t]
\fbox{
\noindent
\hspace{-1.5em}
\begin{minipage}{\columnwidth}
    \begin{algorithmic}
    \small
    \State $\textrm{all\_triple\_sets}\gets \{\}$
    \State $\textrm{rel\_pairs} \gets \{\}$
    \State $\textrm{depth} \gets 5$
    \ForAll{$r_{i} \in KG$}
    \State $P \gets \{(r_j, c_{ij})  \forall (r_i, r_j, c_{ij}) \in \textrm{train\_align\_counts}\}$
    \State $rel\_pairs(r_i) \gets \textrm{maxheap}(P)$ 
    \EndFor
    \ForAll{$\textrm{entities } s \in KG$}
        \State $R \gets \{(r, o) \forall (s, r, o) \in  KG\}$ 
        \While{$R \ne \emptyset$}
            \State $\textrm{triple\_set}\gets \{\}$
            \State $(r_1, o_1) \gets random(R)$
            \State $\textrm{triple\_set}.add(s, r_1, o_1)$
            \State $R.remove(s, r_1, o_1)$
            \State $KG.remove(s, r_1, o_1)$
            \For{$i=2 \textbf{ to } \textrm{depth}$}
                \State $r_i \gets \textrm{NONE}$
                \State $M \gets rel\_pairs(r_{i-1})$
                \While{$M \ne \emptyset$}
                    \State $(r_j, c_{ij}) \gets M.next$
                    \If{$r_j \in R$}
                        \State $r_i \gets r_j$
                        \State $(r_i, o_i) \gets R.get(r_i)$
                        \State $\textrm{triple\_set}.add(s, r_i, o_i)$
                        \State $R.remove(s, r_i, o_i)$
                        \State $KG.remove(s, r_i, o_i)$                       
                        \State \textbf{break}
                    \EndIf
                \EndWhile 
            \EndFor
            \State $\textrm{all\_triple\_sets}.add(\textrm{triple\_set})$
        \EndWhile
    \EndFor
    \end{algorithmic}
\end{minipage}
}
\caption{Entity Subgraph Creation Algorithm using relation co-occurrence counts based on relation--sentence alignment in the training data. Each entity subgraph consists of a maximum of five triples, all with the same subject entity. The first triple is chosen at random. The second triple is chosen such that its relation has the highest co-occurrence count with the relation in the first triple and so on.}
\label{fig:grouping_algo}
\end{figure}

\subsection{Generation}
For each entity subgraph, we concatenate all its triples as before. We perform top 5 sampling with a temperature of 0.5. The bottom 1\% of the generated sentences are filtered out based on the semantic score assigned using the model in \S \ref{sec:filtering}.

\subsection{Human Evaluation}
Generation quality of the \kelm{} Corpus is evaluated using human ratings on a random sample of 200 entity subgraphs. Automatic metrics such as BLEU \cite{papineni-etal-2002-bleu} or BERTscore \cite{zhang2019bertscore} cannot be used due to the lack of gold references. Following prior work, the generated text is rated for two aspects--fluency and semantics, on a scale of 1-5, where 1 means not at all fluent/does not capture meaning at all and 5 means completely fluent/fully captures meaning with no hallucination. We have eight annotators total and each example is rated by two of them. All annotators are linguists, NLP researchers or NLP practitioners and volunteered for the evaluation. We do not use any crowd sourcing platform. For each instance, scores of the two annotators are averaged to get the final rating. The Pearson correlation between the two sets of ratings is 0.56 for semantics and 0.43 for fluency.

\begin{table}[t]
\centering
\small
\setlength{\tabcolsep}{2pt}
\begin{tabularx}{\linewidth}{L{0.2}L{0.25}L{0.25}|C{0.13}C{0.13}|C{0.13}C{0.13}}
\multicolumn{1}{c}{\bf Model} & \multicolumn{1}{c}{\bf Finetuning} & \multicolumn{1}{c|}{\bf Inference} & \multicolumn{2}{c|}{\bf Semantics} & \multicolumn{2}{c}{\bf Fluency}\\
 & \multicolumn{1}{c}{\bf data} & \multicolumn{1}{c|}{\bf data} & mean & stdev & mean & stdev \\\hline
T5-only & WebNLG & Triple & 4.12 & 1.02 & 4.16 & 1.02 \\
T5-only & WebNLG & Subgraph &  3.97 & 1.14 & 4.15 & 0.87 \\
\multicolumn{2}{c}{--- \tekgen{} ---} & Subgraph & 4.36 & 0.87 & 4.60 & 0.58 \\
\end{tabularx} 
\caption{Human evaluation of the generated corpora, on a scale of 1-5, for semantics and fluency.}
\label{table:gen_ablation}
\end{table}

\begin{table*}[t]
\centering
\footnotesize
\setlength{\tabcolsep}{2pt}
\begin{tabularx}{\linewidth}{L{0.54}|L{0.58}|L{0.42}}
\multicolumn{1}{c|}{\bf Input Triples} & \multicolumn{1}{c|}{\bf WebNLG\_Finetuning + Triples\_Inference} & \multicolumn{1}{c}{\bf {\tekgen} + Subgraph\_Inference} \\
\hline

\begin{tabular}{l}(Michelle Obama, height, +71 inch)\end{tabular} & 
\begin{tabular}{l}Michelle Obama's height is +71 inch.\end{tabular} &
\begin{tabular}{l}Michelle Obama is 71 inches tall.\end{tabular} \\
\hline

\begin{tabular}{l}(10x10 Photobooks, instance of, \\ Nonprofit organization), \\ (10x10 Photobooks inception, 00 2012) \end{tabular} & 
\begin{tabular}{l}The 10x10 Photobooks are the result of a \\non-profit organization. \\ 10x10 Photobooks was started in 00 2012.\end{tabular} &
\begin{tabular}{l}10x10 Photobooks, founded in 2012 \\is a nonprofit organization.\end{tabular} \\
\hline

\begin{tabular}{l}(Edu (footballer, born 1949), member of \\sports team, Tigres UANL) \\ (Edu (footballer, born 1949 ), Tigres \\ UANL end time, 01 January 1983) \\ (Edu (footballer, born 1949 ), Tigres \\ UANL start time, 01 January 1978) \end{tabular} &	
\begin{tabular}{l}Edu was born in 1949 and is a member of \\Tigres UANL. \\ Edu ( footballer , born in 1949 ) Tigres \\ UANL's end time was 01 January 1983. \\ Edu ( footballer , born 1949 ) was at Tigres\\ UANL from 01 January 1978. \end{tabular} &
\begin{tabular}{l} Edu, who was born in 1949, played \\for Tigres UANL between 1978 and \\1983.\end{tabular}\\
\hline

\begin{tabular}{l}(To Kill a Mockingbird, award received, \\Pulitzer Prize for Fiction) \\ (To Kill a Mockingbird Pulitzer Prize for \\Fiction point in time 00 1961) \\ (To Kill a Mockingbird Pulitzer Prize for \\Fiction winner Harper Lee) \end{tabular} &
\begin{tabular}{l}To Kill a Mockingbird won the Pulitzer \\ Prize for Fiction. \\To Kill a Mockingbird was Pulitzer Prize \\for Fiction, awarded in 00 1961. \\ Harper Lee was the winner of the Pulitzer \\ Prize for Fiction for To Kill a Mockingbird. \end{tabular} &
\begin{tabular}{l}Harper Lee won the 1961 Pulitzer \\Prize for Fiction for To Kill a \\Mockingbird. \end{tabular} \\
\hline

\begin{tabular}{l}(Caucasus Mountains, country, Georgia \\(country)) \\ (Caucasus Mountains, instance of, \\Mountain range) \\ (Caucasus Mountains, country, Russia) \\ (Caucasus Mountains, highest point, \\Mount Elbrus) \\ (Caucasus Mountains, country, Armenia) \\ \\(Caucasus Mountains, topic's main \\ category, Category:Caucasus Mountains) \end{tabular} &
\begin{tabular}{l}The Caucasus Mountains are located in \\Georgia. \\ The Caucasus Mountains is an example of a \\Mountain range. \\ Caucasus Mountains is in Russia. \\ The highest point in the Caucasus Mountain\\-s is Mount Elbrus. \\ Caucasus Mountains is in the country of \\Armenia. \\ The Caucasus Mountains is categorised as a \\Caucasus Mountains.\end{tabular} &
\begin{tabular}{l}The Caucasus Mountains are a \\mountain range found in Georgia, \\Armenia and Russia. Mount Elbrus \\is the highest point in the Caucasus \\Mountains.\end{tabular}

\end{tabularx} 

\caption{Examples of text generated by the final model in comparison to the model trained only on WebNLG.}
\label{table:gen_ablation_exmples}
\end{table*}

We compare \tekgen{} to two baseline systems. For both baselines, we fine-tune a T5-large model only on WebNLG 2017 data but use different inference input. For one system, we use one triple at a time as input, resulting in 524 instances from the 200 entity subgraphs. For the second, we use the entity subgraphs as input, resulting in 200 instances. Scores are shown in Table \ref{table:gen_ablation}. Entity subgraphs during inference do not improve the mean scores but reduce the standard deviation of the fluency. In comparison, \tekgen{} with inference using entity subgraphs improve both the semantics and fluency of the generated text. Both the mean scores are higher and the standard deviations are lower. It paraphrases canonical names of relations in the KG to more natural expressions more often. Some examples of generation using the two systems are shown in Table \ref{table:gen_ablation_exmples}. In the second example, the relation `inception' is paraphrased to `started' using WebNLG\_finetuning+Triple\_Inference and `founded' using {\tekgen}+Subgraph\_Inference, the latter being more appropriate for organizations.

For completeness, we evaluate two more baseline systems in which T5-large model is finetuned only on the KG--Text aligned corpus but use the two different inference inputs--single triple and entity subgraphs. One annotator rated the same sample for semantics. The former had an average score of 2.34 and the latter 2.73. Since these scores were very low, we did not pursue the evaluation of these systems further. The use of just the aligned corpus which is noisy to some extent results in the worst performing system out of all the methods.

\section{Knowledge Enhanced LMs}

In this section, we showcase an application of the generated \kelm{} Corpus as a way to integrate KGs into natural text corpora for pre-training language models (LMs). 
We choose REALM \cite{guu2020realm} as a representative of the recently introduced family of retrieval language models and therefore we expect our work to be equally applicable to other such language models.
We show gains on LAMA knowledge probe and open domain QA with augmentation.
We also perform experiments where we integrate raw Wikidata triples instead of \kelm{} corpus to confirm the effectiveness of verbalization.

\subsection{Retrieval Language Models}
REALM is a retrieval-based language model and uses two corpora for pre-training--a retrieval corpus and a pre-training corpus. During pre-training, a sentence is selected at random from the pre-training corpus and a random word or salient span (dates and entities) is masked in this sentence. Then using a joint representation of the masked sentence and each of the documents in the retrieval corpus, the masked word is predicted.
In the finetuning stage, the model is provided with a query/question as input in place of masked sentence from the pre-training corpora. It retrieves a small set of documents from the retrieval corpus based on the vector similarity of the query and document representation and then selects a span of text from the retrieved documents as the answer.

\subsection{\kelm{} Documents}
\label{sec:kelm_docs}
We group sentences in the \kelm{} corpus by subject entities to create 5722974 (5.7M) documents.
We call these \kelm{} documents.
We then replace/augment  the retrieval corpus in REALM with these synthetic documents.
\kelm{} Corpus has only $\sim$286M words ($\sim$14\%) in comparison to $\sim$2B words in English Wikipedia.

\begin{table*}[t]
\centering
\small
\setlength{\tabcolsep}{2pt}
\begin{tabularx}{\linewidth}{L{0.64}|C{0.15}C{0.15}C{0.15}C{0.15}|C{0.15}C{0.15}C{0.15}C{0.15}|C{0.15}|C{0.15}}
\textbf{} & \multicolumn{4}{c|}{\bf Google-RE} & \multicolumn{4}{c|}{\bf TREx} & \multirow{2}{*}{\bf Squad} & {\bf Concept}\\
& DOB & POB & POD & Total & 1-1 & N-1 & N-M & Total & & {\bf Net} \\\hline

Elmo 5.5B \cite{peters-etal-2018-deep} & 0.10 & 7.50 & 1.30 & 3.00 & 13.10 & 6.50 & 7.40 & 7.10 & 4.30 & 6.20 \\
Tranformer-XL \cite{dai2019transformer} & 0.90 & 1.10 & 2.70 & 1.60 & 36.50 & 18.00 & 16.50 & 18.30 & 3.90 & 5.70 \\
BERT-large \cite{devlin-etal-2019-bert} & 1.40 & 16.10 & 14.00 & 10.50 & {\bf 74.50} & 34.20 & 24.30 & 32.30 & 17.40 & {\bf 19.20} \\

\hline \hline
\multicolumn{1}{l}{\em REALM} \\
\hline

\textsc{Original} & & & & & & & & & & \\
Wikipedia & 49.06 & 79.56 & 64.13 & 67.36 & 55.81 & 69.54 & 66.98 & 68.18 & 27.96 & 4.78 \\\hline

\textsc{Replaced} & & & & & & & & & & \\
Triple Documents & 69.46 & 61.64 & 53.01 & 63.03 & 49.30 & 62.34 & 53.12 & 58.43 & 18.09 & 4.27 \\
{\kelm} Documents & 68.91 & 61.37 & 53.79 & 62.81 & 49.41 & 61.60 & 52.50 & 57.76 & 19.07 & 4.26 \\\hline

\textsc{Augmented} & & & & & & & & & & \\
Wikipedia + Triple Documents & 71.60 & 80.92 & 69.89 & 76.32 & 57.20 & 69.96 & 67.86 & 68.80 & 29.93 & 4.81 \\
Wikipedia + {\kelm} Documents & {\bf 76.75} & {\bf 83.92} & {\bf 74.86} & {\bf 80.30} & 57.84 & {\bf 70.33} & {\bf 68.09} & {\bf 69.13} & {\bf 31.57} & 5.25 \\
\end{tabularx} 
\caption{Accuracy on LAMA probe. Pretaining corpus is CCnews and the retrieval corpus changed for REALM.}
\label{table:realmLAMA}
\end{table*}

\begin{table}[t]
\centering
\small
\setlength{\tabcolsep}{2pt}
\begin{tabularx}{\linewidth}{L{0.77}C{0.18}C{0.18}}
\textbf{REALM Retrieval Corpus} & \textbf{NQ} & \textbf{WQ} \\\hline

\textsc{Original} \\
Wikipedia (reported) & 40.40 & 40.70 \\
Wikipedia (rerun) & 38.84 & 40.80 \\\hline

\textsc{Replaced} \\
Triple Documents & 21.14 & 42.54 \\
{\kelm} Documents & 22.58 & 41.19 \\\hline

\textsc{Augmented} \\
Wikipedia + Triple Documents & 40.28 & 42.91 \\
Wikipedia + {\kelm} Documents & {\bf 41.47} & {\bf 43.90} \\
\end{tabularx} 
\caption{Exact Match (EM) accuracy of REALM on NQ and WQ. Pretraining corpus used is CCNews.}
\label{table:realmQA}
\end{table}

\subsection{Evaluation Datasets}

We perform evaluation using two open domain question answering datasets and one knowledge probing dataset.

\subsubsection{Question Answering}

\paragraph{NaturalQuestions (NQ)} \cite{47761}: Queries to Google and their answers.

\paragraph{WebQuestions (WQ)} \cite{berant-etal-2010-global}: question-answers from the Google Suggest API.

We keep the same settings as REALM for both NQ and WQ i.e.\ we work on the open domain setting for both datasets where {\em no} passage is provided as context for each question. Finetuning is performed on respective training splits.

\subsubsection{Knowledge Probe}

\paragraph{LAMA} \cite{petroni-etal-2019-language}: Fill-in-the-Blank style factual queries with single token answers from four sources: Google-RE,\footnote{\url{https://code.google.com/archive/p/relation-extraction-corpus/}} T-REx \cite{elsahar-etal-2018-rex}, SQuAD \cite{rajpurkar-etal-2016-squad} and ConceptNet \cite{speer-havasi-2012-representing}. Google-RE also consists of aliases of each answer.

REALM did not include LAMA as one of its evaluation datasets. So we first evaluate REALM on LAMA using the original retrieval corpus and then using the \kelm{} Corpus. No finetuning is performed and the masked word predictions from the pre-trained models are used as answers.

\subsection{Results}
We evaluate REALM on WQ, NQ and LAMA under three settings by modifying the retrieval corpus.

\begin{enumerate}
    \itemsep0em
    \item \textsc{Original}: Wikipedia text
    \item \textsc{Replaced}: only \kelm{} Corpus
    \item \textsc{Augmented}: Wikipedia text + \kelm{} Corpus
\end{enumerate}

The \textsc{Replaced} and \textsc{Augmented} models are evaluated using both the raw Wikidata triples and the generated sentences. Wikidata triples are grouped by subject entity to form Triple Documents and \kelm{} Corpus sentences are also grouped by subject entity to form \kelm{} Corpus Documents (\S \ref{sec:kelm_docs}). The model is pre-trained for 200k steps with the CC-News pre-training corpus in all cases with default hyperparameters.

\paragraph{\textsc{Original}} For NQ and WQ, we fine-tuned the pre-trained REALM on the respective training splits. While we were able to reproduce the accuracy on WQ, the accuracy on NQ was $\sim$1.5\% absolute less than the reported accuracy (row 1\&2 in Table \ref{table:realmQA}). For LAMA probe, we first evaluated the pre-trained REALM, reporting the results on the different sub-corpora in Table \ref{table:realmLAMA} (row {\em Wikipedia} under REALM). Even the \textsc{Original} REALM model shows substantial improvement over prior models. The ability of REALM to access the corpus documents during inference not only make it interpretable but also better on the knowledge intensive tasks. It obtains an accuracy of 67.36\% on Google-RE, 68.18\% on T-REx and 27.96\% on SQuAD. In comparison, the reported accuracy for BERT \cite{devlin-etal-2019-bert} is 10.50\% on Google-RE, 32.30\% on T-REx and 17.40\% on SQuAD. BERT performs better on 1-1 T-REx relations with 74.50\% accuracy as compared to REALM with 55.81\% accuracy. However, this group consists of only two relations; {\em capital} and {\em capital of}. BERT also has better performance than REALM on the ConceptNet subcorpus. On inspection of some of the queries in ConceptNet, we found the questions to be vague and possibly hard for even humans. For example, {\em Raven can \_\_\_} and  {\em Time is \_\_\_}.

\paragraph{\textsc{Replaced}} The \textsc{Replaced} model which uses only \kelm{} Corpus Documents, performs better than the \textsc{Original} model on WQ but the accuracy is much lower on NQ (rows 3\&4 in Table \ref{table:realmQA}). This can be attributed to the nature of the datasets--WQ is a KG-based dataset whereas NQ consists of real queries issued to Google. On LAMA (rows 2\&3 under REALM in Table \ref{table:realmLAMA}), the performance is lower than the \textsc{Original} model but much higher than BERT. Both Triple Documents and \kelm{} Corpus Documents have similar performance. When using just the KG, the format doesn't matter. However, a system trained on raw triples may not generalize for tasks where sentence structure is important.

\paragraph{\textsc{Augmented}} We observe improvements on all the datasets (last two rows of Tables \ref{table:realmLAMA}\&\ref{table:realmQA}) with the \textsc{Augmented} model which uses both the Wikipedia text and the \kelm{} Documents. There is an absolute gain of 2.63\% and 3.10\% on NQ and WQ respectively over the \textsc{Original} model. Similarly, there is an absolute gain of 12.94\%, 0.95\%, 3.61\% and 0.47\% on Google-RE, T-REx, SQuAD and ConceptNet in LAMA respectively. Unlike the \textsc{Replaced} model, the improvement is higher when the generated sentences in \kelm{} Corpus are added instead of the raw Wikidata triples, confirming the effectiveness of verbalization of KG into natural language sentences. Wikipedia is the dominant corpus with 2B words whereas KELM corpus sentences are succinct with a total of 286M words (\S \ref{sec:kelm_docs}) so it is likely the learned representations favour the Wikipedia format which is natural language sentences.

We expect augmenting other retrieval-based models such as DPR \cite{karpukhin2020dense} and RAG \cite{lewis2020retrieval} with the \kelm{} corpus should also improve their performance, given that their enhancements are orthogonal to our contribution.
Moreover, we note that our augmented corpus represents a scalable strategy for future QA systems; by adding only 14\% more tokens to the original REALM model we outperform huge and computationally expensive models such as \cite{roberts-etal-2020-much} (11B parameters) on NQ (35.20 $\rightarrow$ 41.47) and WQ (42.80 $\rightarrow$ 43.90).  Wikipedia is the dominant corpus with 2B words whereas KELM corpus sentences are succinct with a total of 286M words (\S \ref{sec:kelm_docs}) so it is likely the learned representations favour the Wikipedia format which is natural language sentences.

We inspected the errors of the \textsc{Augmented} model with \kelm{} Documents on LAMA. Apart from real errors where the prediction is actually incorrect, there were some false errors that can be broadly classified into three categories--

\begin{enumerate}
    \setlength\itemsep{0em}
    \item Ambiguous Query: e.g.\ In ``X was born in \_\_\_\_'', the answer could be the year or the place of birth but only one of them is acceptable depending on the subcorpus.
    \item Incomplete Answer Set: e.g.\ In ``Konstantin Mereschkowski had a career as \_\_\_\_'', the gold  target is {\em biologist} and the prediction is {\em botanist} but both should be correct.
    \item Answer granularity: The prediction is correct but more specific. e.g.\ In ``On the CPI scale, Kenya ranks \_\_\_\_'', the gold answer is {\em low} but the prediction is {\em 101}, which is in fact correct.
\end{enumerate}

\section{Related Work}
\paragraph{Data-to-Text Generation}
Data-to-Text Generation has several benchmark datasets with slightly different objectives--WebNLG \cite{gardent-etal-2017-webnlg} to convert a group of triples to text, E2ENLG \cite{duvsek2018findings} to convert database key-value pairs or pictures to text, WikiBio \cite{lebret-etal-2016-neural} for biography generation from text, \citet{wiseman-etal-2017-challenges} for text describing score statistics tables of basketball games, both ToTTo \cite{parikh2020totto} and DART \cite{radev2020dart} to generate text given a table and relevant highlighted cells. Many systems \cite{van-der-lee-etal-2018-automated, castro-ferreira-etal-2019-neural, shimorina-gardent-2018-handling} have been developed and evaluated on these datasets, such as graph transformers over structured data \cite{koncel-kedziorski-etal-2019-text}, latent templates for interpretability \cite{wiseman2018learning} and text-to-text generation with T5 \cite{kale2020text}.

\paragraph{KG--Text alignment}
T-REx \cite{elsahar-etal-2018-rex} is a widely used Text--KG aligned corpus, built using systems such as coreference resolution and predicate linkers (details in \S \ref{sec:alignment}).
\citet{logan-etal-2019-baracks} and \citet{chen2020kgpt} also created an aligned corpus using Wikipedia hyperlinks and coreference resolution. (details on comparison in \S \ref{sec:triple_types}). In contrast, we use alias-based heuristics coupled with source text selection constraints to generate a corpus of 16M triples aligned with 8M sentences. Lastly, open information extraction i.e.\ automatic KG construction from text \cite{etzioni2008open, angeli-etal-2015-leveraging, clancy-etal-2019-scalable} inherently create such a corpus but these works generally do not release the extracted KG triples.

\paragraph{Incorporating KGs}
Most prior works on incorporating KG with text often learn KG entity representations and add them to the mention spans linked to the entity \cite{peters-etal-2019-knowledge, yu2020jaket, fevry2020entities} or create subgraphs relevant to the query that are expanded with text in the embedding space \cite{logan-etal-2019-baracks, sun-etal-2019-pullnet, xiong-etal-2019-improving}.
Some others incorporate additional modules. \citet{verga2020facts} extend \citet{fevry2020entities} by adding a triple memory with (subject, relation) encoding as the key and the object encoding as the value. 
\citet{das2017question} use universal schema \cite{riedel-etal-2013-relation} that embeds text and KGs in a shared space for their integration. \citet{k-m-etal-2018-learning} learn a single representation for all the triples mentioned in a sentences during pre-training and update it further in task-specific finetuning.
In contrast, we convert the KG into text and use it to augment the pre-training data.

\section{Future Work}
The \kelm{} corpus sentences covers all facts in the KG but the generated sentences are limited to a given entity and its direct relations to other entities.
For example, given the triples (X, child, Y) and (Y, child, Z), it does not the contain ``Z is a grandchild of X''. More complex sentences could be generated by incorporating multi-hop relations in the KG. 
Recent work has also shown promising results on generating multilingual text from English triples \cite{castro-ferreira-etal-2020-2020,agarwal-etal-2020-machine}.
Our proposed approach can be applied to generate a multilingual corpus of facts in various languages using English Wikidata. 


\section{Conclusion}
In this paper, we converted an entire KG (Wikidata) to natural text (\kelm{} Corpus), tackling various challenges over verbalizing domain-specific benchmark datasets.
We further showcase that KG verbalization can be used to integrate KGs and natural text corpora by including the verbalized KG as additional pre-training data.
We augment a retrieval-based language model with the generated synthetic \kelm{} corpus as a retrieval corpus.
We evaluated the augmented model on open domain QA and a knowledge probe, showing improvements on both. The \kelm{} Corpus is publicly available at \url{https://github.com/google-research-datasets/KELM-corpus}.

\section*{Acknowledgments}
We thank William Woods, Jonni Kanerva, Tania Rojas-Esponda, Jianmo Ni, Aaron Cohen and Itai Rolnick for rating the synthetic corpus sample for human evaluation. We also thank Kelvin Guu for his valuable feedback on the paper.

\bibliography{anthology,custom}
\bibliographystyle{acl_natbib}




\end{document}